\documentclass[journal,comsoc]{IEEEtran}
\usepackage[T1]{fontenc}
\usepackage{cite}

\ifCLASSINFOpdf
   \usepackage[pdftex]{graphicx}
\else
   \usepackage[dvips]{graphicx}
\fi

\usepackage{amsmath}
\usepackage[cmintegrals]{newtxmath}

\usepackage[numbers]{natbib}
\usepackage{booktabs}
\usepackage{multirow}
\usepackage{multicol}
\usepackage{amsmath}
\usepackage{algpseudocode}
\usepackage{mathtools}
\usepackage[linesnumbered,ruled,vlined]{algorithm2e}

\mathchardef\mhyphen="2D

\begin{document}
%
\title{Learning Situation-Conditioned Thinking Policies for Long-Term LLM Agents}

\author{Hong~Su
\IEEEcompsocitemizethanks{\IEEEcompsocthanksitem H. Su is with the School of Computer Science, Chengdu University of Information Technology, Chengdu, China.\\
 E-mail: suguest@126.com. \\
\protect\\
}
\thanks{}}

\markboth{Journal of \LaTeX\ Class Files,~Vol.~14, No.~8, August~2015}%
{Shell \MakeLowercase{\textit{et al.}}: Bare Demo of IEEEtran.cls for IEEE Communications Society Journals}
%

\maketitle

\begin{abstract}
Long-running autonomous agents must reuse accumulated reasoning experience without allowing explicit historical memory and LLM context to grow indefinitely. However, existing memory mechanisms mainly retrieve, summarize, or compress past content and do not directly learn when particular kinds of thinking should be activated or discover new thinking knowledge from temporally dispersed experiences. This paper proposes a situation-conditioned thinking memory framework that transforms historical reasoning experience into a lightweight policy for predicting what should be thought about in the current situation, while leaving detailed reasoning to a large language model. Situations may represent temporal or spatiotemporal evolution rather than only current states. Temporary experiences are also periodically analyzed across multiple independent episodes to identify repeated long-range regularities, which are consolidated into new thinking knowledge and further internalized by the lightweight policy. Experiments show that the learned policy achieves 1.000 F1 on temporal-rule generalization, improves DeepSeek reasoning F1 from 0.789 to 0.868, reduces online processing time from 0.3636 ms to 0.0382 ms per query at 30,000 historical situations, and reaches 1.000 relation-discovery F1 and future-thinking accuracy after sufficient repeated cross-experience evidence.
\end{abstract}


\IEEEpeerreviewmaketitle

\section{Introduction}
\label{sec:introduction}

Long-running large language model (LLM) \cite{zhao2026survey} \cite{naveed2025comprehensive} agents increasingly operate in settings where they must accumulate experience, revisit earlier situations, and reuse previously effective reasoning over extended periods.  This requirement is particularly important for autonomous agents and robots, whose decisions may depend not only on the current observation but also on how a situation has evolved over time.  As interaction continues, however, the amount of historical observations, reasoning traces, outcomes, and reflections grows continuously.  Simply retaining all of this material as explicit memory makes subsequent reasoning increasingly dependent on retrieval, context construction, and repeated LLM processing.  A more scalable agent therefore needs not only to remember past content, but also to progressively internalize reusable knowledge about \emph{what should be thought about} when similar situations reappear.

Existing long-term memory mechanisms mainly address this problem through retrieval, summarization, reflection, or structured memory organization \cite{wang2023augmenting} \cite{wang2025recursively}.  These approaches preserve valuable historical evidence and can improve subsequent reasoning, but they usually continue to depend on an external memory whose size grows with experience.  Summary-based compression can reduce this burden, but may discard temporal ordering and subtle transitions that determine which kind of reasoning is required.  Reflection methods can produce higher-level textual knowledge, yet the resulting reflections are still typically stored and later retrieved rather than learned as an automatic situation-to-thinking policy.  Reasoning distillation provides another alternative, but usually trains a smaller model to reproduce answers or reasoning traces, whereas many long-term agents still benefit from retaining a powerful LLM for detailed reasoning.  In addition, some useful knowledge is not explicitly present in any single historical record: it may only become apparent after multiple temporally separated experiences exhibit a repeated pattern, such as an event $A$ followed much later by $B$ and eventually by $C$.  Therefore, a long-term agent requires a mechanism that can both internalize recurring thinking requirements and discover new reusable thinking knowledge across accumulated experiences.

In this paper, we propose a \emph{situation-conditioned thinking policy} for long-term LLM agents.  Instead of repeatedly retrieving complete historical reasoning traces, the framework learns a lightweight mapping from a situation to the thinking requirement that should be activated under that situation.  A situation may describe a static state, a temporal transition, or a longer spatiotemporal evolution, allowing the same current state to trigger different thinking when it is reached through different histories.  The lightweight model predicts \emph{what or when to think}, while the LLM remains responsible for \emph{how to reason} over the current evidence.  In this way, historical reasoning is used as training material rather than being treated only as permanent online context.

The framework further introduces periodic cross-experience consolidation for knowledge that cannot be extracted reliably from a single record.  Temporally dispersed experiences are first organized into candidate long-range relations, after which repeated support, counterexamples, reverse-order evidence, and cross-period stability are considered during consolidation.  Stable relations are formulated as new thinking knowledge and transformed into additional training material for the lightweight policy.  Consequently, a repeatedly observed pattern such as $A\ldots B\ldots C$ can eventually be internalized as a future thinking rule $A\ldots B \rightarrow$ ``consider possible $C$'' before $C$ is observed.  This mechanism allows the agent to move progressively from explicit historical experience to reusable thinking behavior, while still preserving temporary or episodic memory for unresolved and newly emerging situations.

The main contributions of this work are summarized as follows:
\begin{itemize}
    \item We introduce a situation-conditioned thinking memory mechanism that learns a lightweight policy for predicting what kind of thinking should be activated, while leaving detailed reasoning to the LLM.
    \item We represent situations using temporal or spatiotemporal evolution rather than only the current state, enabling the learned policy to distinguish histories that end in the same observation but require different reasoning.
    \item We develop a periodic cross-experience consolidation mechanism that identifies stable long-range regularities across repeated experiences, formulates them as new thinking knowledge, and converts them into training material for future automatic thinking activation.
\end{itemize}

The remainder of this paper is organized as follows.  Section~\ref{sec:related_work} reviews long-term LLM memory, reflection and experience reuse, and temporal cross-experience knowledge discovery.  Section~\ref{sec:model} presents the proposed situation-conditioned thinking framework and its periodic consolidation mechanism.  Section~\ref{sec:verification} describes the experimental setup and evaluates the proposed method across six complementary experiments.  Finally, Section~\ref{sec:conclusion} concludes the paper and discusses the implications of learning reusable thinking behavior from long-term experience.

\section{Related Work}
\label{sec:related_work}

\subsection{Long-Term Memory and Retrieval for LLM Agents}
\label{subsec:rw_memory}

Long-term memory has become a central component of LLM-based agents because interaction histories can quickly exceed the context window and because useful past experience may need to be reused after long delays. Generative Agents stores natural-language observations, retrieves memories according to relevance, recency, and importance, and periodically synthesizes higher-level reflections to support subsequent planning and behavior~\cite{park2023generative}. MemGPT approaches the same long-context problem from a systems perspective by introducing hierarchical memory tiers and virtual-context management, allowing an agent to move information between limited in-context memory and larger external storage~\cite{packer2023memgpt}. More recent work has further explored structured memory organization. For example, THEANINE links memories according to temporal and cause--effect relations and retrieves memory timelines rather than isolated entries~\cite{ong2025theanine}, while MAGMA represents memories through separate semantic, temporal, causal, and entity graphs and performs policy-guided traversal over these relational views~\cite{jiang2026magma}. These methods substantially improve an agent's ability to preserve and retrieve information over long interaction histories.

The main advantage of retrieval-based memory is that original evidence remains explicitly accessible. Complete histories can therefore preserve detailed information that may be difficult to compress safely, and structured retrieval can expose relations that similarity-only retrieval might overlook. This is especially useful when a past event itself is required as evidence for the current decision. However, explicit memory also creates a scaling dependency: as interaction history grows, the system must continue storing, indexing, ranking, and often inserting historical information into the LLM context. Even when the retrieved subset is small, the online reasoning process remains dependent on an external history and on the quality of the retrieval mechanism. Summary-based compression reduces this burden but can remove temporal ordering or subtle transition information that later becomes important.

Our work addresses a different question from conventional memory retrieval. Instead of asking which historical record should be retrieved, we ask whether repeated historical reasoning can be transformed into a lightweight policy that predicts \emph{what kind of thinking should be activated} in the present situation. Historical records are therefore training material rather than necessarily permanent online context. This distinction also separates our method from structured retrieval approaches such as THEANINE and MAGMA: they organize temporal or causal memories so that an LLM can retrieve them more effectively, whereas our method learns a situation-to-thinking mapping and can consequently operate without an ever-growing online history after consolidation. Explicit memory is still useful for unresolved or newly emerging patterns, but already learned thinking requirements can be internalized.

\subsection{Reflection, Lifelong Experience Reuse, and Reasoning Internalization}
\label{subsec:rw_reflection}

A second line of work improves agents by processing previous interaction outcomes rather than merely retrieving raw observations. Reflexion converts feedback from previous trials into verbal reflections and stores them in episodic memory so that later attempts can avoid earlier mistakes~\cite{shinn2023reflexion}. Generative Agents similarly introduces a reflection process that synthesizes lower-level observations into higher-level statements~\cite{park2023generative}. In embodied lifelong learning, Voyager continually explores Minecraft, refines executable skills using environmental feedback, and stores successful programs in an expanding skill library for later reuse~\cite{wang2023voyager}. These approaches demonstrate that experience becomes more useful when it is transformed into reusable higher-level representations instead of being preserved only as raw trajectories.

Reflection and skill libraries nevertheless retain the learned result mainly as explicit textual memories, plans, or executable behaviors. Their strength is interpretability and flexibility: the LLM can inspect a reflection or retrieve a learned skill and directly reuse or adapt it. Their limitation, relative to the present problem, is that the agent still needs to decide which reflection or skill to retrieve and often invokes the LLM to interpret the retrieved content. Furthermore, reflection commonly improves \emph{what the agent should conclude or do} after a previous trial, whereas our target is narrower and complementary: learning \emph{when a particular type of thinking should be invoked} before detailed reasoning is carried out.

Our approach is also different from reasoning distillation. Distilling Step-by-Step, for example, uses LLM-generated rationales as additional supervision for training a smaller task model, allowing a compact model to reproduce more of the teacher's task-solving capability~\cite{hsieh2023distilling}. Such distillation is valuable when the desired outcome is to replace or approximate the large model. In our framework, the lightweight model is not trained to reproduce the entire rationale or final answer. It learns the mapping from a situation to a compact thinking requirement, while the LLM remains responsible for detailed reasoning over the current evidence. Thus, the division of labor is \emph{small model: what/when to think; large model: how to reason now}. This preserves the flexibility of a general LLM while transferring frequently reusable thinking selection into a cheaper learned policy.

\subsection{Temporal Reasoning and Cross-Experience Knowledge Discovery}
\label{subsec:rw_temporal}

Long-term agent memory increasingly incorporates temporal structure because isolated memory items are insufficient when current meaning depends on how events evolved. THEANINE explicitly links memories through temporal and cause--effect relations and constructs timelines that expose changes across long conversations~\cite{ong2025theanine}. TReMu combines memory with neuro-symbolic temporal reasoning for multi-session dialogue, targeting temporal relations that are difficult to answer from isolated utterances~\cite{ge2025tremu}. Temporal Semantic Memory further distinguishes actual event time from dialogue time and consolidates semantically related information into durative memories that represent persistent or evolving states~\cite{su2026tsm}. These approaches show the benefit of preserving temporal organization rather than treating memory as an unordered collection of semantically similar items.

Recent memory research is also moving beyond storage toward abstraction across trajectories. A recent survey characterizes this evolution as progressing from \emph{Storage}, through \emph{Reflection}, to \emph{Experience}, where cross-trajectory abstraction can produce more general knowledge from multiple interactions~\cite{luo2026storage}. This perspective is closely related to our periodic consolidation mechanism. Nevertheless, detecting a recurring temporal relation is not equivalent to summarizing several records. A relation such as $A\ldots B\leadsto C$ may be absent from every individual record as an explicit statement and become apparent only after multiple temporally separated episodes are compared. Moreover, apparent relations may arise by chance; therefore repeated independent evidence, counterexamples, reverse-order evidence, and temporal stability are important before a candidate is promoted to reusable knowledge.

Our framework combines low-level temporal candidate discovery with LLM-based knowledge consolidation. The low-level stage identifies statistically supported long-range candidates and preserves evidence such as recurrence across independent episodes and enrichment over order-shuffled histories. The LLM then organizes these candidates, compares supporting and contradictory evidence, and formulates stable relations as reusable thinking knowledge. Crucially, the discovered relation is not retained only as another memory item: it is converted into training material for the situation-conditioned thinking policy. Consequently, a repeatedly observed pattern $A\ldots B\ldots C$ can eventually produce a future policy $A\ldots B\rightarrow$ ``consider possible $C$'' before $C$ is observed. This closes the gap between temporal memory organization and learned thinking activation: cross-experience regularities are not only retrieved or described, but progressively internalized into when the agent should think about a particular issue.

\section{Situation-Conditioned Thinking Memory with Cross-Experience Knowledge Discovery}
\label{sec:model}

\subsection{Problem Setting and Overall Principle}
\label{subsec:model_overview}

A long-running autonomous agent repeatedly encounters situations, invokes a large language model (LLM), receives environmental feedback, and accumulates experience.  Directly retaining all previous reasoning traces and repeatedly inserting them into the LLM context is not scalable because explicit history grows with operation time.  The objective of our model is therefore not to memorize every previous reasoning result, but to learn \emph{when a reusable type of thinking should be activated}.  A lightweight policy determines \emph{what should be thought about}, while the LLM remains responsible for detailed reasoning about the current situation.

At interaction step $t$, an experience is represented as
\begin{equation}
 e_t=(s_t,a_t,h_t,y_t,f_t),
 \label{eq:experience}
\end{equation}
where $s_t$ is the observed situation, $a_t$ is the performed action when one exists, $h_t$ is the LLM reasoning or other processing trace, $y_t$ is the observed outcome, and $f_t$ denotes later feedback.  A situation is not restricted to a single instantaneous state.  It may contain a sequence
\begin{equation}
 s_t=\mathcal{S}(x_{t-K+1},\ldots,x_t),
 \label{eq:situation_sequence}
\end{equation}
where $K=1$ represents a static situation and $K>1$ allows temporal, spatial, or spatiotemporal evolution to determine the required thinking.  This distinction is important because two situations can end in the same current state while requiring different reasoning due to their different histories.

The model contains two complementary routes for producing reusable thinking knowledge.  First, a single sufficiently informative experience may directly yield a thinking requirement.  Second, some regularities are invisible in any individual record and emerge only after multiple temporally dispersed experiences are jointly examined.  The second route is essential for long-range patterns such as
\begin{equation}
 A\xrightarrow{\Delta_1}B\xrightarrow{\Delta_2}C,
 \label{eq:long_range_pattern}
\end{equation}
where $A$, $B$, and $C$ may be separated by long and variable intervals.  The model therefore maintains temporary experience records and periodically asks the LLM to discover repeated cross-experience regularities rather than merely summarize them.

\subsection{Direct Thinking Abstraction from Individual Experience}
\label{subsec:direct_abstraction}

For an experience that already contains sufficient evidence, an abstraction operator extracts a reusable thinking requirement
\begin{equation}
 g_i=\mathcal{A}_{\mathrm{LLM}}(e_i),
 \label{eq:direct_thinking_abstraction}
\end{equation}
where $g_i$ describes a reasoning direction rather than a final answer.  Examples include checking a cross-signal inconsistency, analyzing a possible hidden dependency, examining future impact, or seeking additional information.  The corresponding direct training material is
\begin{equation}
 d_i^{\mathrm{dir}}=(s_i,g_i).
 \label{eq:direct_material}
\end{equation}
This material allows the lightweight model to learn that situations with similar structure should activate similar thinking, even when their surface objects are different.

However, Eq.~\eqref{eq:direct_thinking_abstraction} is insufficient when the useful regularity spans multiple experiences.  For example, a high-load event may occur, a weak vibration may appear much later, and a positioning error may emerge after an additional delay.  None of the three records alone reveals the relation.  Treating the records independently would therefore fail to produce the thinking rule that the first two events should trigger consideration of the later failure.

\subsection{Periodic Cross-Experience Knowledge Discovery}
\label{subsec:cross_experience_discovery}

The agent keeps a bounded temporary experience buffer
\begin{equation}
 \mathcal{B}_t=\{e_{t-L+1},\ldots,e_t\},
 \label{eq:temporary_buffer}
\end{equation}
where $L$ denotes the currently retained horizon.  Temporary memory is not intended to become a permanently growing prompt.  Instead, it provides a working set from which previously unknown regularities can be discovered.

A consolidation operation is triggered periodically or when sufficient new evidence has accumulated.  Additional triggers may include repeated unexplained outcomes, persistent prediction errors, or novelty.  We denote the trigger by
\begin{equation}
 \chi_t=\mathbb{I}\!\left[
 \mathrm{period}(t)\vee |\mathcal{B}_t|\geq B
 \vee \mathrm{novelty}_t\vee \mathrm{repeatedFailure}_t
 \right].
 \label{eq:consolidation_trigger}
\end{equation}
When $\chi_t=1$, the LLM performs \emph{cross-experience knowledge discovery}
\begin{equation}
 \mathcal{K}_t=
 \mathcal{D}_{\mathrm{LLM}}(\mathcal{B}_t),
 \label{eq:knowledge_discovery}
\end{equation}
where $\mathcal{D}_{\mathrm{LLM}}$ is explicitly instructed to search for repeated temporal, predictive, relational, or candidate-causal regularities.  Its objective is not to shorten the buffer into a narrative summary.  Instead, it seeks new relations that are not explicitly stated in any individual record.

A discovered knowledge unit is represented as
\begin{equation}
 k_j=(P_j,R_j,\Delta_j,g_j,c_j,E_j),
 \label{eq:knowledge_unit}
\end{equation}
where $P_j$ is the antecedent event/state pattern, $R_j$ is the discovered relation type, $\Delta_j$ records temporal constraints, $g_j$ is the thinking requirement induced by the relation, $c_j$ is its confidence, and $E_j$ is the set of supporting experience identifiers.  For example,
\begin{equation}
 P_j:
 A\xrightarrow{\Delta_1}B\xrightarrow{\Delta_2}C
 \quad\Longrightarrow\quad
 g_j=\text{``consider delayed }C\text{-related risk after }A,B\text{''}.
 \label{eq:pattern_to_thinking}
\end{equation}
The event symbols may have no semantic implication by themselves; evidence for the rule is provided by its repeated occurrence across the history.

The model distinguishes a discovered predictive regularity from a proven causal statement.  A newly induced relation is initially treated as a candidate.  Its confidence can be updated as additional supporting or contradicting experience becomes available,
\begin{equation}
 c_j^{(t+1)}=
 \rho c_j^{(t)}+(1-\rho)\,
 \mathrm{Support}(k_j;\mathcal{B}_{t+1}),
 \label{eq:confidence_update}
\end{equation}
where $0\leq\rho<1$ controls persistence.  Relations that repeatedly predict future observations or survive active validation may be promoted, whereas unsupported relations can be weakened or removed.  This design prevents an LLM-generated hypothesis from being treated automatically as confirmed causality.

\subsection{Conversion of Discovered Knowledge into a Lightweight Thinking Policy}
\label{subsec:policy_learning}

Cross-experience discovery becomes useful for future autonomy only when the discovered knowledge changes subsequent processing.  We therefore convert both direct abstractions and newly discovered regularities into training materials.  Let
\begin{equation}
 \mathcal{D}_t^{\mathrm{dir}}
 =\{(s_i,g_i)\}
 \quad\text{and}\quad
 \mathcal{D}_t^{\mathrm{disc}}
 =\{(P_j,g_j):k_j\in\mathcal{K}_t\}.
 \label{eq:two_material_sources}
\end{equation}
The cumulative learning set is updated by
\begin{equation}
 \mathcal{D}_{t+1}=
 \mathcal{D}_t
 \cup \mathcal{D}_t^{\mathrm{dir}}
 \cup \mathcal{D}_t^{\mathrm{disc}}.
 \label{eq:learning_set_update}
\end{equation}
Thus, LLM consolidation is not the final memory representation.  It produces new learning material that can subsequently be internalized.

A lightweight model $M_{\theta}$ is trained to estimate
\begin{equation}
 M_{\theta}(s)=P_{\theta}(g\mid s),
 \label{eq:thinking_policy}
\end{equation}
where $g$ may be a single thinking activity or a multi-label set.  Sequence models such as Transformers or LSTMs are suitable when order and long-range evolution matter, while simpler machine-learning models can be used when the situation representation is static.  The framework therefore does not require one specific small-model architecture; the key requirement is that the learned carrier preserves the situation-to-thinking relation.

For a newly discovered long-range rule, the training situation need not include the eventual outcome $C$.  Instead, a prefix containing the relevant antecedent pattern, such as $A\ldots B$, can be paired with a future-oriented thinking requirement derived from $k_j$:
\begin{equation}
 (A\ldots B,
 \;g_j=\text{``consider/predict }C\text{-related consequence''}).
 \label{eq:prefix_training}
\end{equation}
Consequently, after learning, the agent can activate the appropriate thinking \emph{before} $C$ actually occurs.

\subsection{Runtime Thinking Activation and Continual Development}
\label{subsec:runtime}

At runtime, the current situation is first processed by the lightweight policy.  The activated thinking set is
\begin{equation}
 \widehat{\mathcal{G}}_t=
 \{g:P_{\theta}(g\mid s_t)\geq\tau\},
 \label{eq:thinking_activation}
\end{equation}
where $\tau$ is an activation threshold.  The LLM then performs current reasoning under these thinking directions,
\begin{equation}
 z_t=
 \mathrm{LLM}(s_t,\widehat{\mathcal{G}}_t),
 \label{eq:guided_reasoning}
\end{equation}
and the resulting action and later outcome create a new experience $e_{t+1}$.  The lightweight model therefore answers \emph{what should be considered now}, whereas the LLM answers \emph{how the current problem should be reasoned through}.

The complete continual-development loop is
\begin{equation}
 \begin{split}
 \text{interaction}
 &\rightarrow \text{temporary experiences}\\
 &\rightarrow \text{cross-experience discovery}\\
 &\rightarrow \text{new thinking knowledge}\\
 &\rightarrow \text{policy learning}\\
& \rightarrow \text{future thinking activation}\\
 &\rightarrow \text{new interaction and validation}.
 \end{split}
 \label{eq:continual_loop}
\end{equation}
This loop allows the system to acquire knowledge that was not predefined and was not evident in a single observation.  Repeated experiences can reveal a previously unknown long-range relation; the relation can then be converted into a thinking requirement; and the requirement can finally become an automatically activated lightweight policy.  Exact episodic records may still be retained when needed for auditing or factual recall, but reusable thinking knowledge no longer requires replaying the entire growing history to the LLM.

\section{Verification}
\label{sec:verification}

\subsection{Experimental Setting, Scenarios, and Metrics}
\label{subsec:verification_setup}

We evaluate whether long-running experience can be transformed into reusable thinking knowledge rather than remaining as an ever-growing collection of explicit historical records.  The controlled setting emulates an autonomous robot that continuously observes several heterogeneous signals, including temperature, current, vibration, object distance, visibility, motion state, and load.  A situation is represented by either a single observation or a sequence of observations, so that the experiments can distinguish a current state from the temporal or spatial evolution that produced it.  The experiments are intentionally synthetic and controlled: the objective is not to reproduce a particular robot platform, but to isolate the memory and thinking mechanisms while keeping the correct thinking requirements unambiguous.

The lightweight situation-conditioned thinking policy is implemented by a Transformer encoder with four encoder layers, model dimension 192, six attention heads, a feed-forward dimension of 768, and dropout of 0.1.  It is trained as a multi-label predictor using binary cross-entropy and AdamW.  Five independent random seeds are used for the learning experiments.  The downstream large language model (LLM) is the online \texttt{deepseek-v4-flash} model.  The lightweight model does not generate the final natural-language reasoning result.  Instead, it predicts which relation or issue should be considered, and DeepSeek performs the detailed reasoning over the current situation.

Six experiments test complementary mechanisms.  Experiment~1 tests whether a learned policy avoids the storage and online-retrieval growth of explicit long-term history.  Experiment~2 tests transfer to unknown objects when the required thinking depends on a hidden temporal-order rule.  Experiment~3 isolates the need to represent situation evolution by giving different histories with the same final state.  Experiment~4 tests whether learned thinking guidance improves actual DeepSeek reasoning.  Experiment~5 measures how a rare thinking pattern is acquired as historical experience accumulates.  Experiment~6 tests the newly introduced periodic cross-experience knowledge-discovery mechanism: temporally separated events must recur across multiple independent episodes before they can be consolidated into new knowledge and then internalized as a future thinking policy.

The compared methods are selected according to the mechanism tested.  \emph{Summary-RAG} retrieves historical situations represented by aggregate statistics and is therefore compact but insensitive to temporal order.  \emph{Sequence-RAG} stores the complete temporal sequence and provides a stronger retrieval reference.  \emph{RF-Summary} is a random-forest multi-label classifier trained on the same aggregate features.  \emph{Current-State-Only} uses only the last observation rather than the complete situation evolution.  For the downstream LLM test, \emph{DeepSeek-Only} receives the current observations without learned guidance, \emph{RAG+DeepSeek} receives thinking guidance from retrieved histories, \emph{Proposed+DeepSeek} receives guidance predicted by the learned policy, and \emph{Oracle+DeepSeek} receives the ground-truth thinking requirement.  In Experiment~6, \emph{No Consolidation} does not form cross-experience knowledge, \emph{Unordered} destroys the long-range order, \emph{Local} preserves only short-range order, \emph{Statistics-Only} ranks long-range candidates using their accumulated temporal evidence without LLM consolidation, and \emph{Oracle} directly uses the hidden regularities.

Thinking prediction is evaluated with micro-precision, micro-recall, and micro-F1.  End-to-end LLM reasoning is measured using factor precision, factor recall, factor F1, and final decision accuracy.  Scalability is measured by stored bytes, online historical records, estimated history-context tokens, and measured retrieval or policy-inference time.  Experiment~6 additionally reports candidate recall, relation-discovery F1, and future-thinking activation accuracy.  Unless otherwise stated, values are means and standard deviations over five seeds.

\subsection{Experiment 1: Long-History Consolidation and Scalability}
\label{subsec:exp1}

The first experiment asks whether reusable thinking knowledge can be consolidated into a fixed policy instead of repeatedly searching a growing explicit history.  The history is increased from 500 to 30,000 situations.  Summary-RAG and Sequence-RAG retrieve the top four records for every new query.  Summary-RAG stores order-insensitive aggregate features, whereas Sequence-RAG stores the complete temporal sequence.  The proposed method learns from accumulated histories and subsequently uses no historical record in its online context.

Table~\ref{tab:exp1} shows that Sequence-RAG and the proposed method both maintain a micro-F1 of 1.000 at all four history sizes.  Hence, the scalability result is not obtained by deliberately weakening retrieval accuracy.  Sequence-RAG storage, however, grows from 0.156~MB at 500 histories to 9.360~MB at 30,000 histories, whereas the proposed model remains approximately 7.358~MB.  From the measured per-record index size, the Sequence-RAG index reaches the learned-model size at approximately $2.36\times10^{4}$ historical situations.  Thus, explicit retrieval is initially more compact for a small history, but its memory grows approximately linearly while the learned policy has a fixed parameter size after consolidation.

\begin{figure}[t]
\centering
\includegraphics[width=\linewidth]{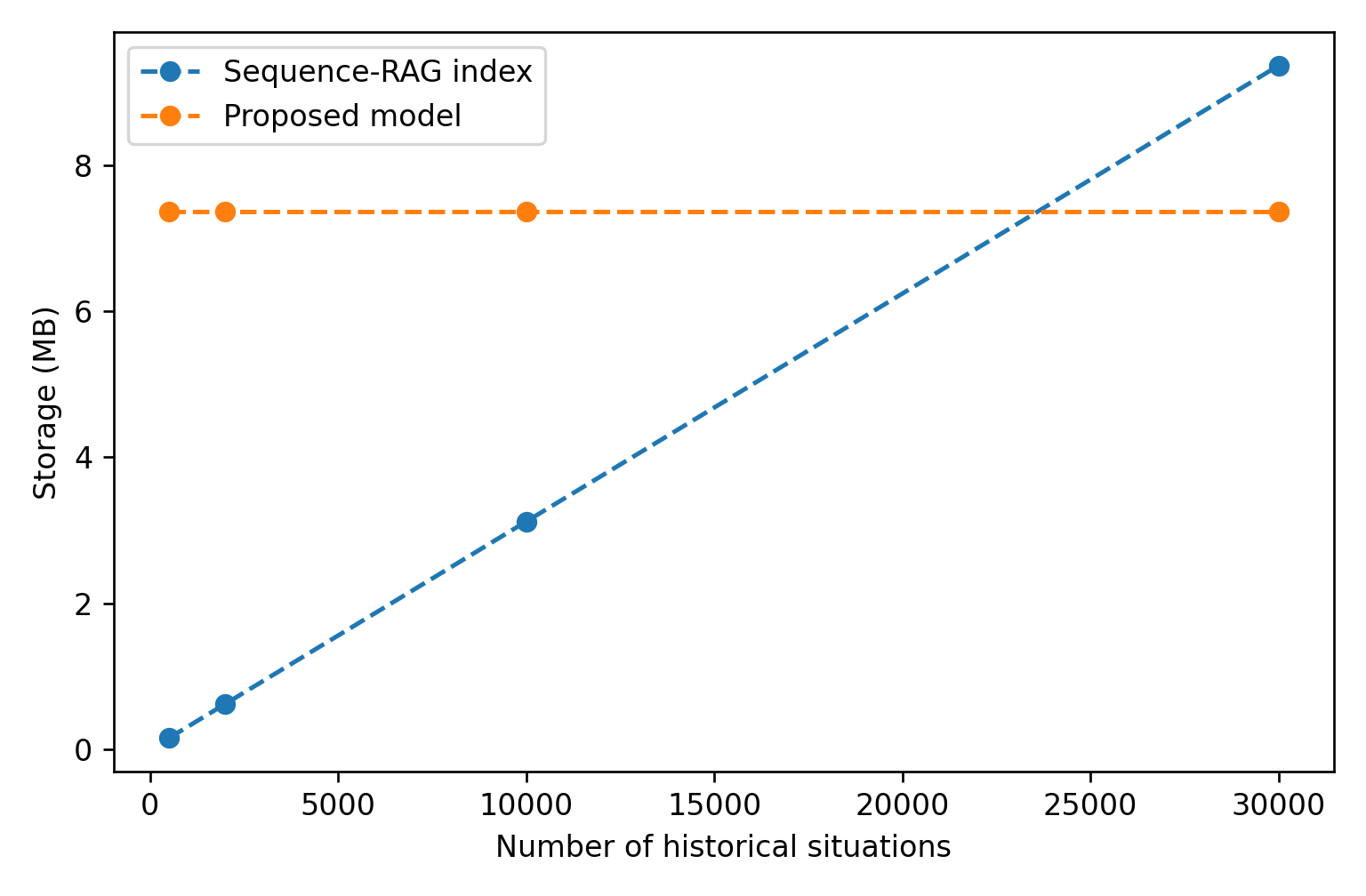}
\caption{Storage as explicit historical experience grows.  Sequence-RAG grows with the number of stored histories, while the consolidated thinking policy remains fixed.}
\label{fig:exp1_storage}
\end{figure}

\begin{figure}[t]
\centering
\includegraphics[width=\linewidth]{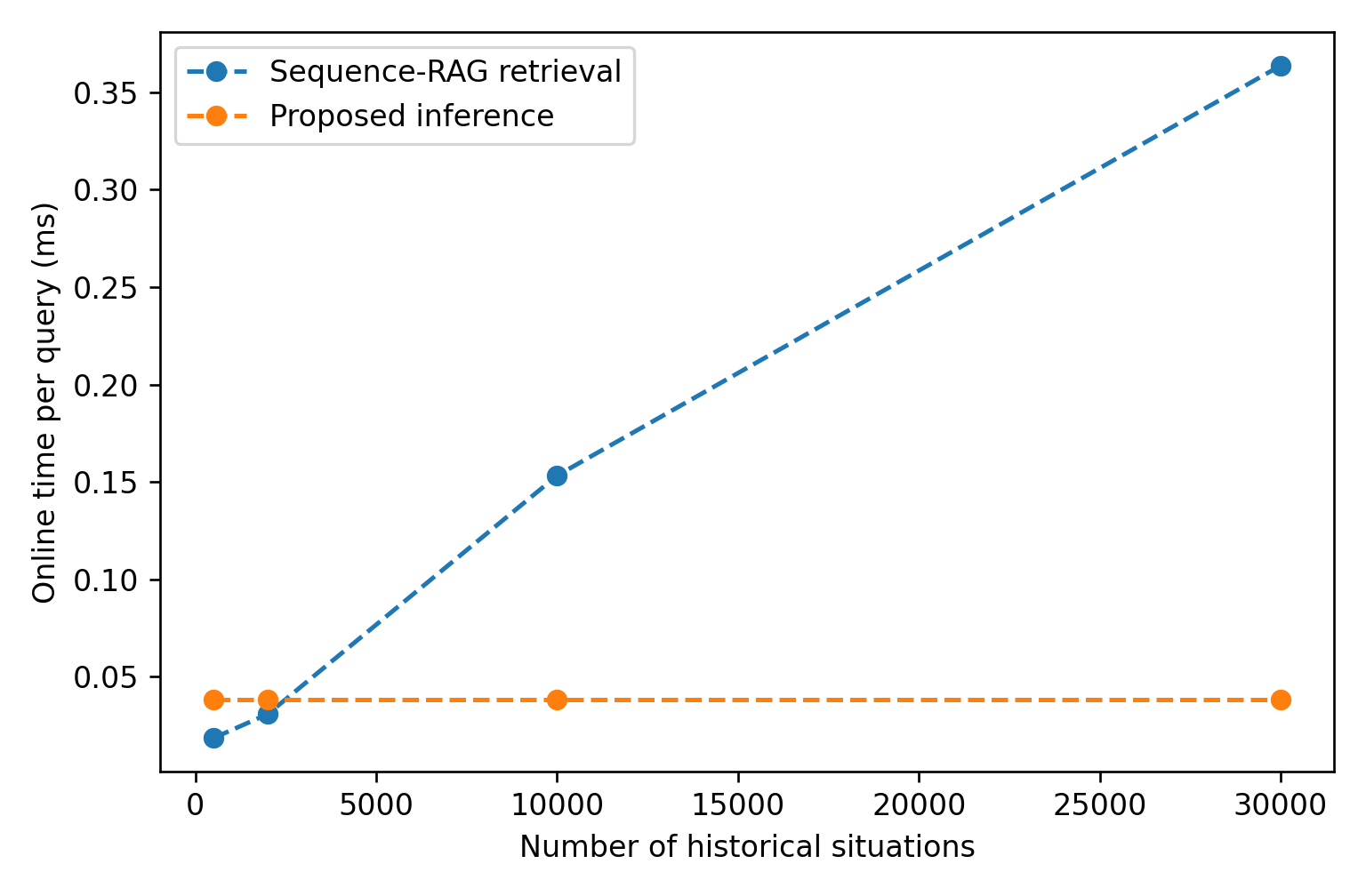}
\caption{Online cost under growing history size.  Sequence-RAG must search a larger explicit memory, whereas the learned policy performs fixed-size inference.}
\label{fig:exp1_latency}
\end{figure}

The online-time trend is similar.  Sequence-RAG retrieval rises from 0.0186~ms/query at 500 histories to 0.3636~ms/query at 30,000 histories, whereas policy inference remains approximately 0.0382~ms/query.  At 30,000 histories, the measured Sequence-RAG retrieval time is therefore about $9.53\times$ the policy-inference time.  Summary-RAG uses a smaller representation but attains only $0.495\pm0.071$ F1 at 30,000 histories because the aggregate representation discards temporal order.  Figures~\ref{fig:exp1_storage} and~\ref{fig:exp1_latency} visualize these two scalability effects.  The experiment therefore demonstrates the intended trade-off: complete retrieval can preserve accuracy but incurs a history-dependent storage and online-search cost; simple history compression is cheaper but can lose temporally structured thinking information; learned consolidation retains the reusable thinking behavior with zero online historical records.

\begin{table*}[t]
\centering
\caption{Long-history scalability with top-$4$ retrieval. Storage uses decimal MB ($10^6$ bytes).}
\label{tab:exp1}
\begin{tabular}{rccccccc}
\hline
History $N$ &
Summary-RAG F1 &
Sequence-RAG F1 &
Seq.-RAG MB &
Seq.-RAG ms/query &
Proposed F1 &
Proposed MB &
Proposed ms/query \\
\hline
500    & $0.528\pm0.068$ & $1.000\pm0.000$ & 0.156 & 0.0186 & $1.000\pm0.000$ & 7.358 & 0.0382 \\
2,000  & $0.518\pm0.042$ & $1.000\pm0.000$ & 0.624 & 0.0309 & $1.000\pm0.000$ & 7.358 & 0.0382 \\
10,000 & $0.499\pm0.073$ & $1.000\pm0.000$ & 3.120 & 0.1533 & $1.000\pm0.000$ & 7.358 & 0.0382 \\
30,000 & $0.495\pm0.071$ & $1.000\pm0.000$ & 9.360 & 0.3636 & $1.000\pm0.000$ & 7.358 & 0.0382 \\
\hline
\end{tabular}
\end{table*}

\subsection{Experiment 2: Generalization of Non-Obvious Temporal Thinking Rules}
\label{subsec:exp2}

Experiment~2 tests whether the learned memory captures a reusable thinking rule rather than recognizing an object name or an obvious physical relation.  Training situations contain one set of robot-component identities, while testing situations contain unseen identities.  More importantly, different classes are constructed to have the same start value, final value, set of intermediate values, mean, standard deviation, and global change.  The classes differ only in the \emph{ordering} of their intermediate observations, and the mapping from each ordering to its required thinking activity is an environment-specific rule rather than common world knowledge.

Table~\ref{tab:exp23} shows that the proposed sequence Transformer obtains $1.000\pm0.000$ micro-F1.  Summary-RAG reaches $0.566\pm0.045$, while RF-Summary reaches $0.383\pm0.033$.  Because aggregate sequence statistics were deliberately matched, the approximately 0.434 F1 gap between Proposed and Summary-RAG cannot be attributed to a different final state or a different mean value; it arises from preserving and learning the temporal ordering itself.  The result therefore supports the claim that the learned thinking memory can transfer an environment-specific situation-to-thinking relation to unknown objects when the same latent evolution reappears.

\subsection{Experiment 3: Same Final State, Different Situation Evolution}
\label{subsec:exp3}

The third experiment isolates the importance of situation evolution even further.  Each pair ends at exactly the same moderate temperature and current values.  In one history, however, the signals rise toward the endpoint and require trend, cause, and hazard analysis; in the other, they fall toward the same endpoint and indicate recovery.  A current-state-only model therefore sees identical final evidence for two cases that should activate different thinking.

The proposed sequence-aware policy achieves $1.000\pm0.000$ micro-F1, compared with $0.711\pm0.044$ for Current-State-Only (Table~\ref{tab:exp23}).  This is an absolute difference of approximately 0.289.  The result demonstrates that ``what should be thought about'' may depend on how the present state was reached, not merely on the current observation itself.  Together, Experiments~2 and~3 justify representing a situation as a temporal or spatiotemporal sequence when the relevant thinking requirement is transition-dependent.

\begin{table}[t]
\centering
\caption{Mechanism verification for temporal-order generalization and situation evolution.}
\label{tab:exp23}
\begin{tabular}{llc}
\hline
Experiment & Method & Micro-F1 \\
\hline
Exp.~2 & Proposed Transformer & $1.000\pm0.000$ \\
       & Summary-RAG & $0.566\pm0.045$ \\
       & RF-Summary & $0.383\pm0.033$ \\
\hline
Exp.~3 & Proposed Sequence & $1.000\pm0.000$ \\
       & Current-State-Only & $0.711\pm0.044$ \\
\hline
\end{tabular}
\end{table}

\subsection{Experiment 4: End-to-End Thinking Guidance with DeepSeek}
\label{subsec:exp4}

Experiment~4 tests whether the learned thinking policy improves downstream LLM reasoning rather than only reproducing thinking labels.  Every test situation contains several simultaneous numerical signal sequences.  The lightweight policy outputs a reasoning direction, such as comparing temperature with current, examining visibility jointly with distance, comparing vibration with load, or inspecting a late current change.  These directions are \emph{not} the final answers.  DeepSeek must still inspect the current numerical observations and determine which canonical factor is actually present.  Test objects use unseen identities so that the result cannot be obtained from a memorized component name.

As shown in Table~\ref{tab:exp4}, DeepSeek-Only obtains a factor F1 of $0.789\pm0.038$.  Proposed+DeepSeek increases this value to $0.868\pm0.035$, an absolute increase of 0.079 and approximately a 10.0\% relative improvement.  Recall is already 1.000 for all four methods, whereas factor precision increases from $0.704\pm0.055$ to $0.815\pm0.048$.  The improvement therefore mainly reflects fewer irrelevant or incorrect factors, i.e., the learned policy helps the LLM focus on the relevant relation rather than simply encouraging it to produce more candidate factors.  Decision accuracy also increases from $0.484\pm0.115$ to $0.575\pm0.140$.

RAG+DeepSeek and Oracle+DeepSeek both obtain $0.877\pm0.029$ factor F1, only 0.009 above Proposed+DeepSeek.  A paired analysis over the 320 saved situation-level comparisons gives a Proposed--DeepSeek factor-F1 difference of $+0.0786$ with 95\% bootstrap interval $[0.0609,0.0974]$ and paired randomization $p<0.001$.  Decision correctness improves by $+0.0906$, with interval $[0.0594,0.1250]$ and $p<0.001$.  In contrast, the Proposed--RAG factor-F1 difference is $-0.0089$ with interval $[-0.0302,0.0135]$ and $p=0.472$, so there is no reliable accuracy difference between these two methods in this controlled task.  This result is consistent with the intended role of consolidation: Proposed approaches the quality of explicit retrieval while Experiment~1 shows that it does not require a growing online history.

\begin{table*}[t]
\centering
\caption{End-to-end reasoning with \texttt{deepseek-v4-flash}. Values are mean$\pm$standard deviation over five seeds.}
\label{tab:exp4}
\begin{tabular}{lccccc}
\hline
Method & Factor Precision & Factor Recall & Factor F1 & Decision Accuracy & Avg. Tokens \\
\hline
DeepSeek-Only &
$0.704\pm0.055$ & $1.000\pm0.000$ & $0.789\pm0.038$ & $0.484\pm0.115$ & $719.61\pm0.98$ \\
RAG+DeepSeek &
$0.835\pm0.039$ & $1.000\pm0.000$ & $0.877\pm0.029$ & $0.559\pm0.106$ & $727.94\pm0.82$ \\
Proposed+DeepSeek &
$0.815\pm0.048$ & $1.000\pm0.000$ & $0.868\pm0.035$ & $0.575\pm0.140$ & $728.13\pm1.18$ \\
Oracle+DeepSeek &
$0.835\pm0.039$ & $1.000\pm0.000$ & $0.877\pm0.029$ & $0.559\pm0.106$ & $727.94\pm0.82$ \\
\hline
\end{tabular}
\end{table*}

\begin{table*}[t]
\centering
\caption{Paired per-situation analysis in Experiment~4 ($n=320$).}
\label{tab:exp4_stats}
\begin{tabular}{lccc}
\hline
Comparison / Metric & Mean Diff. & 95\% CI & $p$ \\
\hline
Proposed -- DeepSeek / F1 & $+0.0786$ & $[0.0609,0.0974]$ & $<0.001$ \\
Proposed -- DeepSeek / Decision & $+0.0906$ & $[0.0594,0.1250]$ & $<0.001$ \\
Proposed -- RAG / F1 & $-0.0089$ & $[-0.0302,0.0135]$ & $0.472$ \\
Proposed -- RAG / Decision & $+0.0156$ & $[-0.0281,0.0594]$ & $0.587$ \\
\hline
\end{tabular}
\end{table*}

\begin{figure}[t]
\centering
\includegraphics[width=\linewidth]{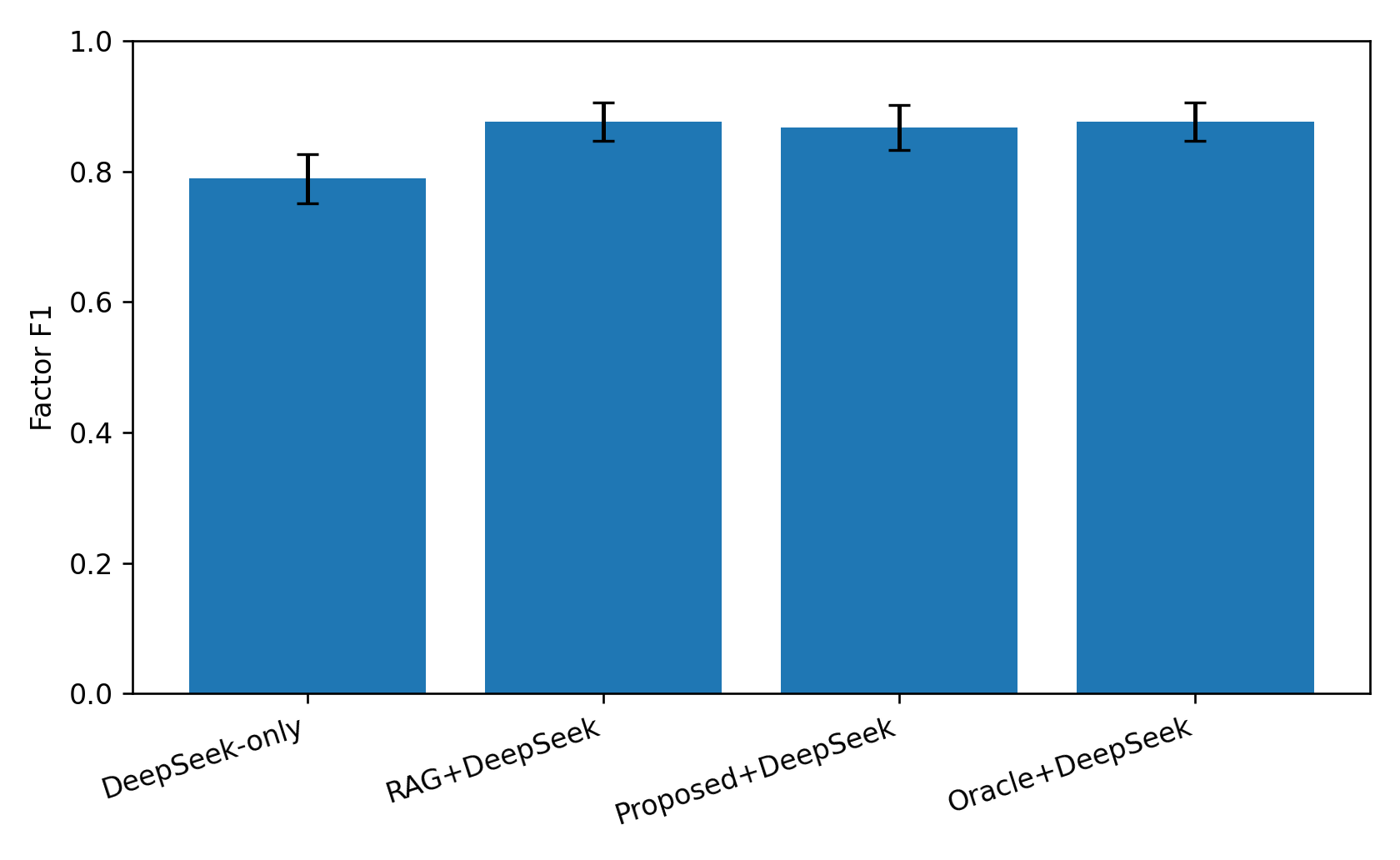}
\caption{Factor F1 in the end-to-end DeepSeek experiment.  Learned thinking guidance improves over unguided DeepSeek and approaches the explicit-retrieval and oracle references.}
\label{fig:exp4}
\end{figure}

\subsection{Experiment 5: Formation of Thinking Knowledge from Rare Experience}
\label{subsec:exp5}

Experiment~5 tests whether a thinking policy can actually emerge from accumulated experience.  A delayed-onset event is made rare in the historical stream: signals remain apparently normal for a long prefix and then exhibit a late abnormal acceleration.  Correct handling requires trend, hazard, cause, future-impact, and information-seeking thinking.  The test set contains this rare pattern, while the number of available historical situations is increased from 100 to 8,000.

\begin{figure}[t]
\centering
\includegraphics[width=\linewidth]{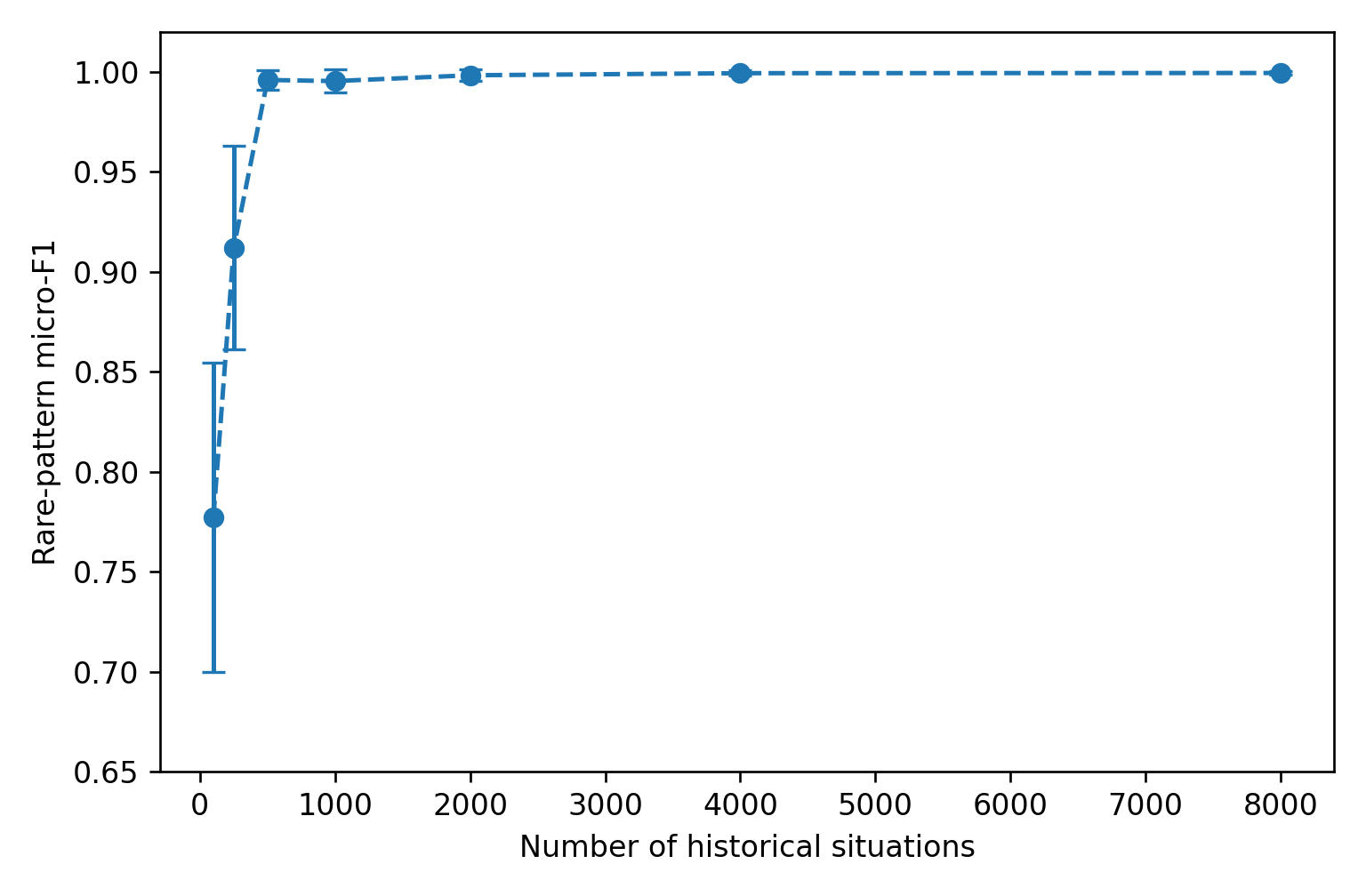}
\caption{Learning of a rare thinking pattern as the number of historical situations increases.  Error bars show one standard deviation over five seeds.}
\label{fig:exp5}
\end{figure}

Figure~\ref{fig:exp5} shows a clear acquisition curve.  With 100 histories, rare-pattern micro-F1 is $0.777\pm0.077$.  It increases to $0.912\pm0.051$ at 250 histories and $0.996\pm0.005$ at 500.  Performance then approaches saturation, reaching $0.998\pm0.003$ at 2,000 histories and $0.999\pm0.001$ at 8,000.  Thus, the thinking policy is not assumed to exist from the beginning: it becomes reliable after enough relevant historical evidence has accumulated, with diminishing returns once the rare pattern has been internalized.

\begin{table}[t]
\centering
\caption{Rare-pattern learning as historical experience accumulates.}
\label{tab:exp5}
\begin{tabular}{rc}
\hline
Historical situations & Proposed micro-F1 \\
\hline
100   & $0.777\pm0.077$ \\
250   & $0.912\pm0.051$ \\
500   & $0.996\pm0.005$ \\
1,000 & $0.995\pm0.006$ \\
2,000 & $0.998\pm0.003$ \\
4,000 & $0.999\pm0.001$ \\
8,000 & $0.999\pm0.001$ \\
\hline
\end{tabular}
\end{table}

\subsection{Experiment 6: Periodic Cross-Experience Discovery of New Thinking Knowledge}
\label{subsec:exp6}

The final experiment evaluates the new periodic cross-experience consolidation mechanism.  Its purpose is different from Experiments~2--5: the target thinking rule is \emph{not explicitly available in any individual record}.  Each seed uses randomly generated event codes with no semantic meaning, such as $E_{016}$, $E_{176}$, and $E_{030}$.  A hidden regularity has the form
\[
A \;\cdots\; B \;\cdots\; C,
\]
where $A$, $B$, and $C$ are separated by long random intervals.  The same relation must recur across multiple independent episodes before it should be regarded as reusable knowledge.  Since the event codes are randomly assigned, DeepSeek cannot infer the answer from pretrained common knowledge.

The history also contains deliberately difficult counterevidence: isolated occurrences of $A$, $B$, and $C$; $A\cdots B$ episodes in which $C$ does not follow; reversed $B\cdots A$ sequences; neutral episodes; and decoy triples that repeat strongly in only one temporal partition.  A low-level temporal stage therefore produces an over-complete set of candidate relations and measures support across independent episodes, counterexamples, reverse evidence, cross-partition stability, and enrichment relative to event-order shuffling.  This stage does not declare the final knowledge.  DeepSeek periodically compares the accumulated evidence and converts accepted relations into explicit reusable thinking knowledge.  The accepted relation is then materialized as training data so that a lightweight policy can activate ``think about possible $C$'' when a future $A\cdots B$ prefix reappears before $C$ occurs.

Table~\ref{tab:exp6} shows a pronounced evidence-accumulation threshold.  With only one or two positive episodes per hidden rule, candidate recall and Proposed relation F1 are both 0.  At four repetitions, candidate recall rises to $0.933\pm0.149$, but Proposed relation F1 is only $0.333\pm0.471$.  Thus, most true rules have begun to enter the candidate set, but the accumulated evidence is not yet sufficient for stable consolidation across seeds.  At eight independent repetitions, candidate recall reaches $1.000\pm0.000$, and Proposed+DeepSeek reaches $1.000\pm0.000$ relation F1.  This transition supports a three-stage interpretation: insufficient evidence, candidate emergence, and stable knowledge consolidation.

The ablations at eight repetitions further isolate what makes the knowledge discoverable.  No Consolidation, Unordered, and Local all obtain 0 relation F1, whereas Statistics-Only, Proposed+DeepSeek, and Oracle all obtain 1.000.  Destroying long-range order or restricting the analysis to local structure therefore removes the signal, while repeated long-range evidence is sufficient to identify the hidden relation.  The fact that Statistics-Only matches Proposed in this controlled synthetic task is important: it indicates that the LLM is not required to outperform statistical evidence ranking for numerical relation identification.  Its role is instead to organize the accumulated cross-experience evidence, formulate accepted relations as explicit reusable knowledge, and convert them into new thinking materials.

\begin{figure}[t]
\centering
\includegraphics[width=\linewidth]{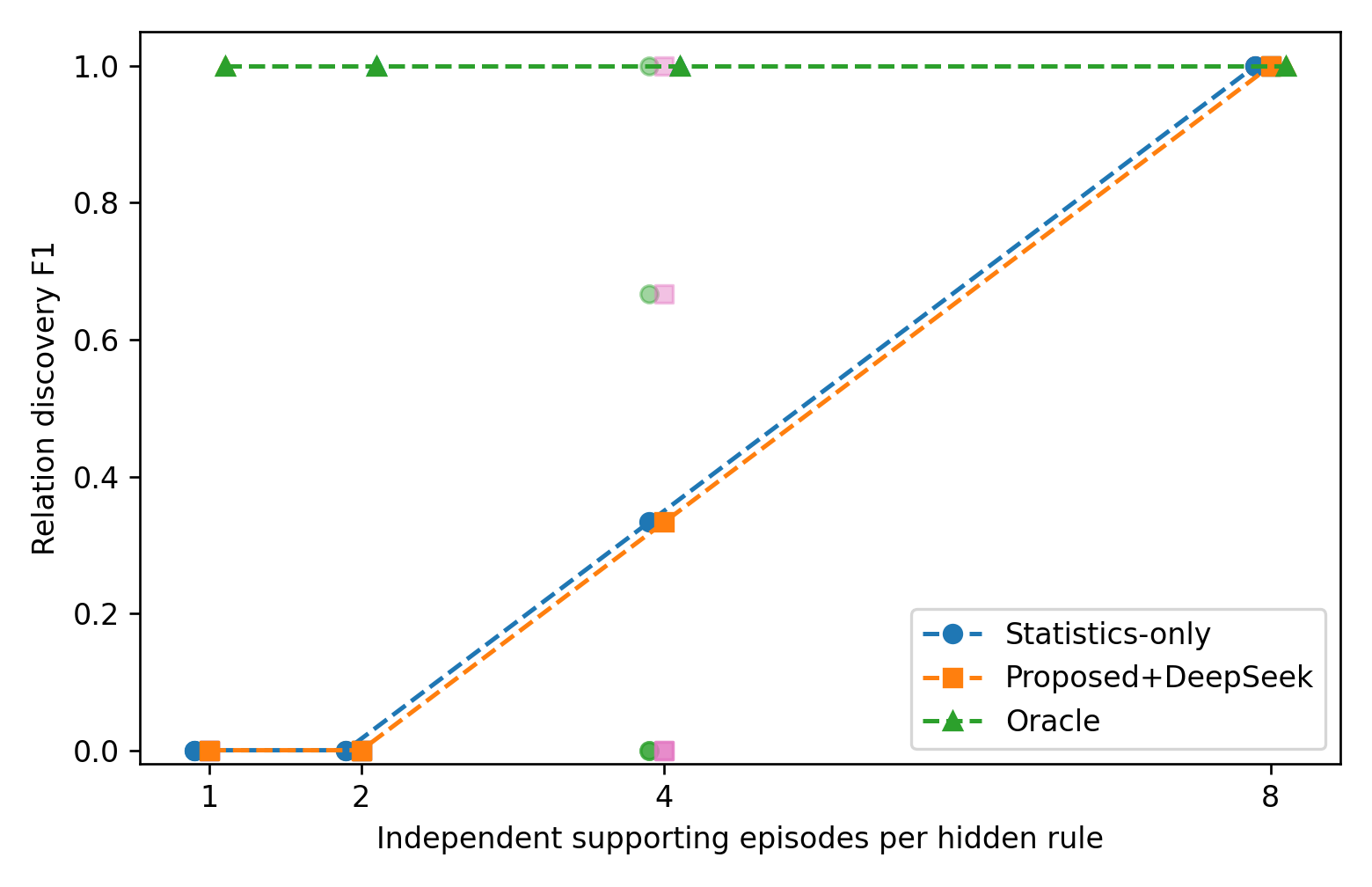}
\caption{Cross-experience relation discovery as the number of independent
supporting episodes increases. Markers with lower opacity show individual
random-seed results, while dashed lines show the mean performance. The large
variation at four repetitions indicates an intermediate stage in which true
relations have begun to emerge but are not yet consolidated consistently
across runs.}
\label{fig:exp6}
\end{figure}

This distinction is confirmed by the downstream internalization test.  At eight repetitions, the future-thinking policy trained from Statistics-Only, Proposed, or Oracle relations reaches an accuracy of $1.000\pm0.000$, whereas No Consolidation, Unordered, and Local remain at $0.333\pm0.000$, which is the chance level for three possible future-thinking targets.  Hence, once a stable $A\cdots B\leadsto C$ regularity has been consolidated, it can be transformed into a policy that activates the appropriate thinking about $C$ before $C$ is observed.  Figure~\ref{fig:exp6} summarizes the relation-discovery transition as repeated independent evidence accumulates.

\begin{table*}[t]
\centering
\caption{Cross-experience knowledge discovery.  Candidate recall and relation F1 are reported over five seeds.  Future-thinking accuracy is evaluated at eight repetitions.}
\label{tab:exp6}
\begin{tabular}{rcccc}
\hline
Repetitions/rule &
Candidate Recall &
Statistics-Only F1 &
Proposed+DeepSeek F1 &
Oracle F1 \\
\hline
1 & $0.000\pm0.000$ & $0.000\pm0.000$ & $0.000\pm0.000$ & $1.000\pm0.000$ \\
2 & $0.000\pm0.000$ & $0.000\pm0.000$ & $0.000\pm0.000$ & $1.000\pm0.000$ \\
4 & $0.933\pm0.149$ & $0.333\pm0.471$ & $0.333\pm0.471$ & $1.000\pm0.000$ \\
8 & $1.000\pm0.000$ & $1.000\pm0.000$ & $1.000\pm0.000$ & $1.000\pm0.000$ \\
\hline
\multicolumn{5}{l}{\footnotesize At 8 repetitions, No-Consolidation, Unordered, and Local relation F1 are all $0.000\pm0.000$.}
\end{tabular}
\end{table*}

\begin{table}[t]
\centering
\caption{Future-thinking activation after cross-experience consolidation at eight repetitions per hidden rule.}
\label{tab:exp6_future}
\begin{tabular}{lc}
\hline
Knowledge source & Future-thinking accuracy \\
\hline
No Consolidation & $0.333\pm0.000$ \\
Unordered & $0.333\pm0.000$ \\
Local & $0.333\pm0.000$ \\
Statistics-Only & $1.000\pm0.000$ \\
Proposed+DeepSeek & $1.000\pm0.000$ \\
Oracle & $1.000\pm0.000$ \\
\hline
\end{tabular}
\end{table}

Overall, the six experiments form a connected verification chain.  Experiment~1 shows why reusable thinking knowledge should be consolidated rather than indefinitely retrieved from a growing explicit history.  Experiments~2 and~3 show that the policy can preserve non-obvious and transition-dependent thinking requirements.  Experiment~4 demonstrates that the learned guidance improves actual DeepSeek reasoning.  Experiment~5 shows that a thinking policy emerges as relevant experience accumulates.  Experiment~6 extends this process beyond single experiences: repeated temporally dispersed events can first be recognized as a stable cross-experience regularity, converted into new thinking knowledge, and finally internalized so that the corresponding thinking is activated before the future outcome occurs.

\section{Conclusion}
\label{sec:conclusion}

This paper proposed a situation-conditioned thinking memory framework for long-term LLM agents. Instead of repeatedly retrieving an ever-growing reasoning history, the framework learns a lightweight policy that predicts what should be thought about under the current situation, while leaving detailed reasoning to the LLM. The model further represents situations as temporal or spatiotemporal evolutions and periodically consolidates temporally dispersed experiences to identify stable cross-experience regularities, transform them into new thinking knowledge, and internalize them for future activation. Experimental results show that the learned policy preserves non-obvious temporal thinking rules, improves DeepSeek factor F1 from 0.789 to 0.868, maintains fixed-size online reasoning as explicit history grows, and reduces retrieval time by about 9.53$\times$ at 30,000 historical situations. Cross-experience experiments further show that sufficiently repeated long-range evidence can be consolidated into reusable knowledge and subsequently activate correct future thinking before the predicted outcome occurs. These results indicate that long-term agent memory can move beyond storing or retrieving past content toward progressively learning what to think about from accumulated experience.


\ifCLASSOPTIONcaptionsoff
  \newpage
\fi

\bibliographystyle{IEEEtran}
\bibliography{ref}

%

\begin{IEEEbiography}{Hong Su}
  received the MS and PhD degrees, in 2006 and 2022, respectively, from Sichuan University, Chengdu, China. He is currently a researcher of Chengdu University of Information Technology Chengdu, China. His research interests include blockchain, cross-chain and smart contract.
\end{IEEEbiography}




\end{document}